\documentclass[conference]{IEEEtran}
\IEEEoverridecommandlockouts
\usepackage{cite}
\usepackage{amsmath,amssymb,amsfonts}
\usepackage{algorithmic}
\usepackage{graphicx}
\usepackage{textcomp}
\usepackage{xcolor}
\usepackage{comment}
\usepackage{url}
\usepackage[T1]{fontenc}
\usepackage[caption=false,font=footnotesize]{subfig}
\usepackage{placeins}
\usepackage{float}
\usepackage{todonotes} %TODO comment out for final version

\def\BibTeX{{\rm B\kern-.05em{\sc i\kern-.025em b}\kern-.08em
    T\kern-.1667em\lower.7ex\hbox{E}\kern-.125emX}}
\begin{document}

\title{DINO-Explorer: Active Underwater\\ Discovery via Ego-Motion Compensated \\Semantic Predictive Coding
}
\author{
\IEEEauthorblockN{
Yuhan Jin\IEEEauthorrefmark{1},
Nayari Marie Lessa\IEEEauthorrefmark{1}\IEEEauthorrefmark{2},
Mariela De Lucas Alvarez\IEEEauthorrefmark{1},
Melvin Laux\IEEEauthorrefmark{1}\IEEEauthorrefmark{2},
Lucas Amparo Barbosa\IEEEauthorrefmark{3},\\
Frank Kirchner\IEEEauthorrefmark{1}\IEEEauthorrefmark{2}, and
Rebecca Adam\IEEEauthorrefmark{1}
}
\IEEEauthorblockA{\IEEEauthorrefmark{1}%
Robotics Innovation Center, German Research Center for Artificial Intelligence, 
Bremen, Germany\\
\IEEEauthorblockA{\IEEEauthorrefmark{2}
Robotics Research Group,
University of Bremen,
Bremen, Germany}
\IEEEauthorblockA{\IEEEauthorrefmark{3}
Computing Department,
SENAI CIMATEC,
Salvador, Brazil}
Correspondence: yuhan.jin@dfki.de
}

%\author{\IEEEauthorblockN{Yuhan Jin}
%\and
%\and
%\IEEEauthorblockN{Rebecca Adam}
%\IEEEauthorblockA{\textit{Robotics Innovation Center} \\
%\textit{
%German Research Center for Artificial Intelligence
%}\\
%Bremen, Germany \\
%https://orcid.org/0000-0003-1080-1330}
%\and
%\and
%\and
}
\maketitle

\begin{abstract}
Marine ecosystem degradation necessitates continuous, scientifically selective underwater monitoring. However, most autonomous underwater vehicles (AUVs) operate as passive data loggers, capturing exhaustive video for offline review and frequently missing transient events of high scientific value. Transitioning to active perception requires a causal, online signal that highlights significant phenomena while suppressing maneuver-induced visual changes. We propose DINO-Explorer, a novelty-aware perception framework driven by a continuous semantic surprise signal. Operating within the latent space of a frozen DINOv3 foundation model, it leverages a lightweight, action-conditioned recurrent predictor to anticipate short-horizon semantic evolution. An efference-copy-inspired module utilizes globally pooled optical flow to discount self-induced visual changes without suppressing genuine environmental novelty. We evaluate this signal on the downstream task of asynchronous event triage under variant telemetry constraints. Results demonstrate that DINO-Explorer provides a robust, bandwidth-efficient attention mechanism. At a fixed operating point, the system retains 78.8\% of post-discovery human-reviewer consensus events with a 56.8\% trigger confirmation rate, effectively surfacing mission-relevant phenomena. Crucially, ego-motion conditioning suppresses 45.5\% of false positives relative to an uncompensated surprise signal baseline. In a replay-side Pareto ablation study, DINO-Explorer robustly dominates the validated peak F1 versus telemetry bandwidth frontier, reducing telemetry bandwidth by 48.2\% at the selected operating point while maintaining a 62.2\% peak F1 score, successfully concentrating data transmission around human-verified novelty events.

\end{abstract}
\section{Introduction}
\begin{figure}[ht]
\centering
\includegraphics[width=\columnwidth]{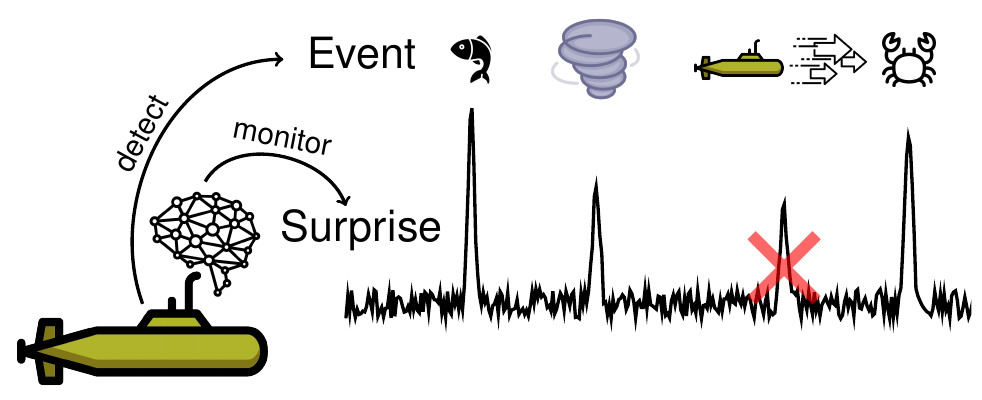}
\caption{Conceptual overview of DINO-Explorer: Inspired by predictive coding, the framework generates an intrinsic surprise signal that acts as an indicator of semantic novelty to serve active discovery. Downstream tasks, such as detecting scientifically relevant events (e.g., biological anomalies or habitat transitions), can then be driven by this signal. Furthermore, to mitigate false positives caused by the AUV's own motion, the system isolates genuine environmental changes by implementing an efference copy-inspired ego-motion compensation module.}
\label{fig:teaser}
\end{figure}

Underwater robots are increasingly used for ecosystem observation, scientific survey, and offshore infrastructure inspection, but the resulting visual streams remain costly to review frame by frame \cite{petillot2019underwater, gonzalez2023survey}. Remotely Operated Vehicles (ROVs) and Autonomous Underwater Vehicles (AUVs) can now collect underwater imagery at scales that are impossible to inspect manually \cite{petillot2019underwater}. At the same time, underwater vision remains fundamentally difficult: light scattering, color distortion, turbidity, and marine snow all degrade the sensory stream seen by the robot \cite{gonzalez2023survey}. These effects are especially severe in the shallow North Sea, where tides, waves, and storms continually resuspend sand and mud from the seabed \cite{opdal2019centennial, wilson2019increasing}.

This creates a perception bottleneck for curiosity-driven underwater robots \cite{zhou2022discovering}. A robust novelty indicator for this setting must do more than record long missions: it needs to highlight transient, semantically meaningful changes despite turbidity, illumination drift, marine snow, and frequent camera motion, while remaining stable enough for downstream robot interfaces. Three technical obstacles make this difficult. First, low-level motion estimators based on optical flow \cite{teed2020raft} or visual odometry \cite{ferrera2019real} are effective for ego-motion estimation, but their geometric and photometric cues are too brittle to define semantic novelty in visually degraded water. Second, a mobile platform continuously generates self-induced visual change, so any novelty signal must separate externally caused events from the sensory consequences of the AUV's own maneuvers. Third, the signal must remain causal and compact enough to be reused for downstream robot decisions rather than optimized only for an offline review dataset.

We address these obstacles by defining \textit{surprise} in semantic latent space. DINO-Explorer draws on two classic ideas from neuroscience and physiology: predictive coding \cite{Rao1999, friston2005theory}, which interprets surprise as deviation from an expected sensory state, and efference copy \cite{vonHolst1950, crapse2008corollary}, which explains how self-generated sensory consequences can be discounted. In that sense, DINO-Explorer is a bio-inspired robotics framework: it turns these  principles into a practical attention mechanism for an embodied underwater robot. We instantiate these ideas with frozen DINOv3 latent states \cite{simeoni2025dinov3}, short-horizon semantic prediction, and ego-motion cues to build a causal surprise signal for underwater robot attention allocation. This signal is not tied to a single downstream task: in robotics it can support selective review, adaptive sensing, and later closed-loop decision modules. This paper evaluates one representative downstream interface: asynchronous underwater event triage under mission-time telemetry constraints.

Specifically, we investigate three core questions: First, can semantic prediction mismatch identify mission-relevant semantic changes despite severe underwater visual noise? Second, can motion conditioning prevent the robot from mistaking its own maneuvers for genuine environmental novelty? Third, when used as a causal trigger policy, can the same surprise signal filter long survey missions and reduce telemetry bandwidth while retaining reviewer-agreed events? We answer these questions by evaluating the downstream trigger policy on offline replay data and a selected operating point against human consensus.

In summary, our primary contributions are threefold:

\begin{enumerate}
    \item \textbf{Motion-aware semantic surprise modeling:} We formulate a lightweight predictive baseline in frozen DINOv3 latent space and use one-step semantic prediction error to define a continuous surprise signal that is more robust than pixel-space novelty cues in noisy underwater video.
    \item \textbf{Ego-motion-aware false-alarm suppression:} We condition the predictive model on globally pooled optical flow, giving the system an efference-copy-style motion cue that discounts maneuver-induced appearance change instead of treating it as external novelty.
    \item \textbf{Validation of surprise signal with downstream triage interface:} We treat the continuous surprise score as the core output of the method and evaluate a representative downstream triage interface using a Pareto sweep on replay data, a fixed protocol for human verification, and an analysis of telemetry reduction. The sweep tests the full event-proposal-quality/bandwidth trade-off instead of a single selected threshold; the consensus summary verifies that the selected operating point retains events broadly identified by the human annotators.
\end{enumerate}

\section{Related Work}

Relevant prior work spans three main areas: active perception and underwater selective discovery, foundation-model representations for semantic perception, and predictive state-space modeling in robotics. Classical active-perception research asks what an embodied agent should sense and when it should spend attention or computation \cite{bajcsy1988active, bajcsy2018revisiting}. Most underwater systems, by contrast, still emphasize robust acquisition, mapping, or task-specific inspection rather than online semantic triage \cite{petillot2019underwater, gonzalez2023survey}. A notable published exception is context-enhanced anomaly modeling for curiosity-driven underwater exploration \cite{zhou2022discovering}, which targets underwater anomaly-driven discovery more directly than full survey-stack autonomy pipelines.

Foundation-model representations have recently strengthened underwater semantic perception. DINOv2 and DINOv3  features provide a strong semantic substrate for marine downstream tasks \cite{chen2026autonomous} \cite{simeoni2025dinov3} . In underwater settings, published studies already show value for expert-assisted marine image annotation protocols \cite{orenstein2025assisting} and downstream tasks such as instance segmentation with frozen DINO backbones \cite{chen2026empowering}. Together, these results suggest that foundation-model state spaces can remain useful even under severe marine-domain degradation.

Outside the underwater domain, related robotics and world-model literature connects semantic prediction, intrinsic motivation, and self-motion-aware sensing. Curiosity-driven exploration via self-supervised prediction links semantic novelty signals to intrinsic-motivation work \cite{pathak17a}, while world-model lines from cognitive robotics to latent-dynamics planning and physical robot learning provide a broader conceptual lineage for expectation-based novelty and self-motion-aware sensing \cite{taniguchi2023world, ha2018worldmodels, hafner2019planet, wu2023daydreamer}. Recent foundation-model world models such as DINO-WM extend this line to pretrained DINO features by learning dynamics for planning without pixel reconstruction \cite{zhou2025dinowm}. Practical robotic deployments ground those ideas in deployable motion cues through estimators such as Recurrent All-Pairs Field Transforms (RAFT) and underwater visual odometry \cite{teed2020raft, ferrera2019real}. Taken together, these lines motivate continuous surprise not as a task-locked anomaly score, but as a reusable robot-facing signal for attention allocation, selective communication, and later control interfaces. DINO-Explorer evaluates one underwater triage instantiation of that broader robotics role.

\begin{figure*}[!h]
\centerline{\includegraphics[width=\textwidth]{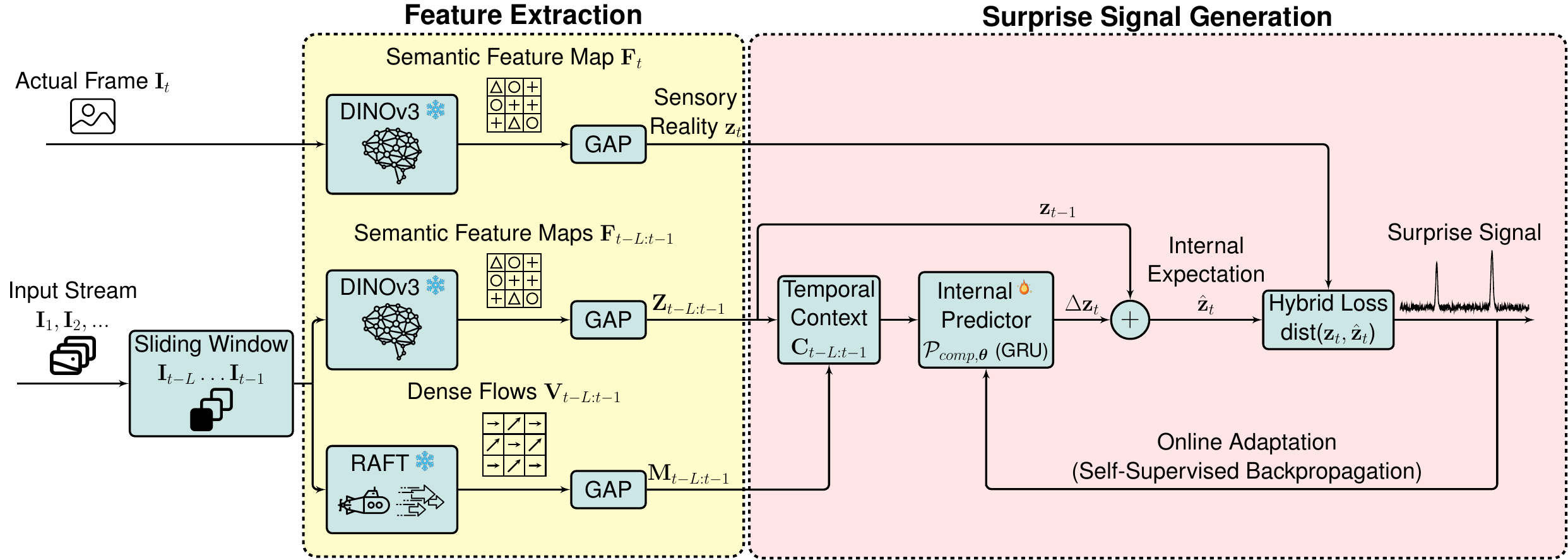}}
\caption{System architecture of DINO-Explorer: The pipeline comprises \textit{Feature Extraction} (left/yellow), which produces actual semantic state $\mathbf{z}_t$, history semantic states $\mathbf{Z}_{t-L:t-1}$ and RAFT-based motion estimates $\mathbf{M}_{t-L:t-1}$, and \textit{Surprise Signal Generation} (right/pink). Within a predictive coding framework, the GRU-based ego motion compensated semantic predictor ($\mathcal{P}_{comp, \boldsymbol{\theta}}$) generates an internal expectation $\hat{\mathbf{z}}_t$ conditioned on the motion-aware temporal context $\mathbf{C}_{t-L:t-1}$, which acts as an efference copy. The resulting surprise signal $\mathcal{S}_t$, quantifying the mismatch between sensory reality $\mathbf{z}_t$ and internal expectation $\hat{\mathbf{z}}_t$, is used both to drive continuous online adaptation via self-supervised backpropagation and to provide a real-time novelty indicator for downstream active discovery tasks.}
\label{fig:system_diagram}
\end{figure*}

\section{Preliminary}

\subsection{Predictive Coding and Curiosity-Driven Learning}
Predictive coding provides the conceptual lens for expectation-based perception: an agent maintains an internal prediction of the next sensory state, and surprise arises when observation departs from that expectation \cite{Rao1999, friston2005theory}. In curiosity-driven robot learning, the same mismatch can be reused as an intrinsic signal that highlights informative states instead of relying on raw appearance change alone \cite{pathak17a, taniguchi2023world}. For robot systems, that kind of surprise signal can gate attention, sampling, or later action selection without committing the representation to a single downstream task.

\subsection{Ego-Motion Compensation and Efference Copy}
Efference copy provides the complementary lens for discounting self-generated sensory change \cite{vonHolst1950, crapse2008corollary}. For embodied robots, this principle motivates conditioning semantic expectations on a compact estimate of self-motion, so maneuver-induced transients can be treated as expected reafferent change rather than externally caused novelty. In underwater settings, where direct motor telemetry is often limited or unreliable, optical-flow-based motion cues offer one practical route to that conditioning \cite{teed2020raft}.

Together, these two principles motivate a robot-facing continuous surprise signal: a predictor models expected semantic evolution, and a motion cue discounts robot-induced appearance change.

\section{DINO-Explorer}
The Dino-Explorer framework comprises a frozen semantic observation model, a recurrent transition model, an optical-flow-derived action-conditioning module, and a causal event extractor, serving as a downstream task example. The first three components define the internal world model and produce the continuous surprise signal; the last converts that signal into mission-facing review proposals.

Figure~\ref{fig:system_diagram} presents the Dino-Explorer system architecture, which consists of two primary components. The yellow block represents \textit{Feature Extraction}, which generates the semantic state representation. The red block denotes \textit{Surprise Signal Generation}, where semantic states are processed to produce \textit{Surprise signals} when instances deviate from predicted regularities, as well as the self-supervised backpropagation signal for online module correction.

Unless noted otherwise, in this section, bold lowercase letters denote vectors, bold uppercase letters denote matrices, tensors, or vector sequences, uppercase Roman letters denote fixed dimensions, window lengths, counts, or aggregate energies, and lowercase Roman letters denote indices or per-step scalars. Greek letters denote scalar hyperparameters, and calligraphic letters denote predictors or continuous signals such as $\mathcal{P}$ and $\mathcal{S}$.

\subsection{Semantic State Representation}
To strictly focus on global scene semantics, our architecture operates exclusively on a 1D global vector level, as shown in the overall pipeline (Fig.~\ref{fig:system_diagram}). Let $\mathbf{I}_t \in \mathbb{R}^{H \times W \times 3}$ denote the input video frame at time step $t \in \{1, \dots, T\}$. To overcome the stochasticity of pixel-level underwater noise, we utilize the DINOv3 vision foundation model, denoted as a frozen encoder $f_\phi$, to map input frames to a robust semantic latent space. For an input image $\mathbf{I}_t$, the encoder produces a dense feature map $\mathbf{F}_t \in \mathbb{R}^{H' \times W' \times D}$:

\begin{equation}
    \mathbf{F}_t = f_\phi(\mathbf{I}_t).
\end{equation}
where $H'=32, W'=32$ are the spatial dimensions of the patch tokens for an input resized to $512 \times 512$ pixels, and $D$ is the feature dimension (e.g., $D=1024$ for ViT-L). To obtain a compact global state representation $\mathbf{z}_t \in \mathbb{R}^D$, we apply global average pooling across the spatial dimensions:
\begin{equation}
    \mathbf{z}_t = \frac{1}{H' \times W'} \sum_{i=1}^{H'} \sum_{j=1}^{W'} \mathbf{F}_t^{(i,j)}.
\end{equation}
This operation collapses the spatial dimensions into a single 1D vector $\mathbf{z}_t$, effectively abstracting away localized artifacts such as caustic lighting or marine snow to focus on global scene characteristics.

\begin{figure*}[htbp]
\centerline{\includegraphics[width=\textwidth]{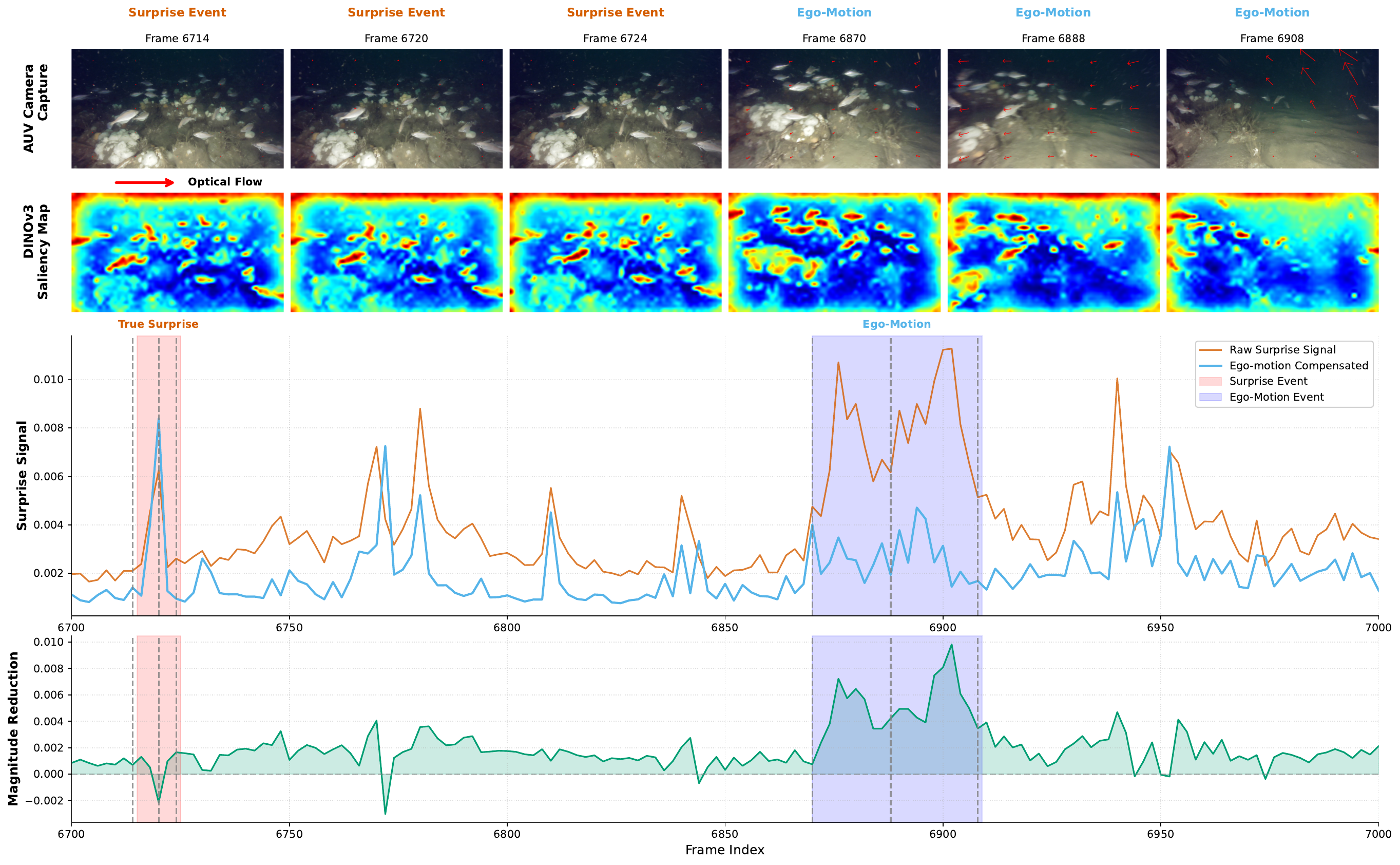}}
\caption{Qualitative analysis of the semantic surprise signal across a continuous sequence. The plot illustrates the efficacy of our predictive coding framework in isolating genuine novelty: (left/red) DINO-Explorer detects significant deviations from the predicted semantic path caused by a fish rapidly changing its trajectory, generating a valid surprise peak that is correctly preserved after ego-motion compensation; and (right/blue) the efference copy-inspired module effectively filters out predictable reafferent noise induced by an AUV maneuver, suppressing a ``naive'' false alarm to ensure the autonomous trigger remains robustly focused on scientifically relevant phenomena.}
\label{fig:surprise_signal_demo}
\end{figure*}

\subsection{Predictive Modeling and Naive Surprise Generation}
DINO-Explorer uses a Gated Recurrent Unit (GRU) predictor $\mathcal{P}_{\boldsymbol{\theta}}$ as a lightweight internal world model parameterised by $\theta$. Given recent semantic states, it predicts the next latent transition one step ahead. The predictor is first pre-trained on continuous underwater footage so it can internalize typical marine dynamics before online adaptation.

The temporal context is represented as a history buffer $\mathbf{Z}_{t-L:t-1} = [\mathbf{z}_{t-L}, \dots, \mathbf{z}_{t-1}]$ with a lookback window $L=50$. At each time step $t$, a GRU processes this buffer via a residual learning formulation to predict the semantic change:

\begin{align}
\Delta \hat{\mathbf{z}}_t = \mathcal{P}_{\boldsymbol{\theta}}(\mathbf{Z}_{t-L:t-1}),
\end{align}
with the predicted semantic state defined as:
\begin{equation}
    \hat{\mathbf{z}}_t = \mathbf{z}_{t-1} + \Delta \hat{\mathbf{z}}_t.
\end{equation}

We quantify the discrepancy between the observed state $\mathbf{z}_t$ and the predicted state $\hat{\mathbf{z}}_t$ using an intrinsic surprise signal as $\mathcal{S}_t$. To capture both magnitude and directional shifts in the latent space, $\mathcal{S}_t$ is calculated with a hybrid loss combining Mean Squared Error (MSE) and Cosine Similarity:
\begin{align}
    \mathcal{S}_t = \|\mathbf{z}_t - \hat{\mathbf{z}}_t\|_2^2 + \lambda (1 - \text{sim}(\mathbf{z}_t, \hat{\mathbf{z}}_t)),
    \label{eq:hybrid_loss}
\end{align}
where $\lambda=0.5$ balances the two components.

%This surprise signal fulfills two roles here.
This surprise signal serves two purposes in this context.
First, it serves as an online indicator of environmental novelty, flagging instances in which the scene deviates from predicted regularities (e.g., the sudden appearance of a benthic habitat). Second, it drives continuous self-supervised adaptation. During deployment, the model backpropagates $\mathcal{S}_t$ at every step to refine its weights $\boldsymbol{\theta}$ using a learning rate $\eta=0.001$:
\begin{align}
\boldsymbol{\theta}_{t+1} = \boldsymbol{\theta}_t - \eta \nabla_{\boldsymbol{\theta}} \mathcal{S}_t(\mathbf{z}_t, \hat{\mathbf{z}}_t(\boldsymbol{\theta}_t)).
\end{align}
This iterative update enables the internal world model to track gradual contextual drift while maintaining high sensitivity to transient anomalies. %As shown in Fig.~\ref{fig:surprise_signal_demo} (left/red),
Therefore, this mechanism effectively isolates novelty, such as the sudden trajectory change of a fish, by identifying significant deviations from the predicted semantic path.

\subsection{Ego Motion Compensation with Optical Flow}
\label{sec:method_egomotion}
To mitigate false positives induced by vehicle maneuvers, we incorporate an action conditioning module grounded in \textit{efference copy}. Since precise motor telemetry is often unavailable or drift-prone in underwater settings, we approximate the agent's motor state using dense optical flow $\mathbf{V}_t$ computed via a RAFT estimator~\cite{teed2020raft}. We derive a scale-invariant 2D global translation vector $\mathbf{m}_t = [\bar{v}_{x}/W, \bar{v}_{y}/H] \in \mathbb{R}^2$ by spatially averaging the $x$ and $y$ components of the flow field (denoted as $\bar{v}_x$ and $\bar{v}_y$) and normalizing the result by the image width $W$ and height $H$.

This global pooling mechanism ensures that the motion cue primarily captures the overall background shift caused by the vehicle rather than the localized movement of individual objects. While moving fauna generate isolated flow vectors, the background, which dominates the frame, exhibits a consistent global shift corresponding to the AUV's trajectory. Averaging the flow across the frame yields a robust proxy for ego-motion with reduced sensitivity to transient subject dynamics.

We formulate the ego motion compensated predictor $\mathcal{P}_{comp, \boldsymbol{\theta}}$ as a conditional GRU. At each time step $t$ within the lookback window $L$, the global semantic state $\mathbf{z}_{t} \in \mathbb{R}^D$ is concatenated with the 2D global motion vector $\mathbf{m}_{t} \in \mathbb{R}^2$ to form $\mathbf{c}_{t} = [\mathbf{z}_{t}, \mathbf{m}_{t}] \in \mathbb{R}^{D+2}$. The temporal sequence $\mathbf{C}_{t-L:t-1} = [\mathbf{c}_{t-L}, \dots, \mathbf{c}_{t-1}]$ is processed by a two-layer GRU with hidden dimension 256, allowing the predictor to model how vehicle motion shifts the high-level semantic state.

The final hidden state is mapped back into the semantic space by a fully connected linear layer, yielding the motion-compensated predicted semantic change:
\begin{align}
\Delta \hat{\mathbf{z}}_{comp, t} = \mathcal{P}_{comp, \boldsymbol{\theta}}(\mathbf{C}_{t-L:t-1}).
\end{align}
The compensated internal expectation $\hat{\mathbf{z}}_{comp, t}$ is subsequently formed by applying this anticipated shift to the prior semantic state $\mathbf{z}_{t-1}$:
\begin{equation}
    \hat{\mathbf{z}}_{comp, t} = \mathbf{z}_{t-1} + \Delta \hat{\mathbf{z}}_{comp, t}.
\end{equation}
% Conditioning on the motion sequence $\mathbf{m}_i$ preserves the original semantic output space while allowing the predictor to associate AUV maneuvers with the resulting visual shifts. The compensated expectation $\hat{\mathbf{z}}_{comp, t}$ can then be compared directly with the incoming semantic state $\mathbf{z}_t$ to isolate externally caused novelty.

Following this motion-conditioned prediction, the framework proceeds identically to the uncompensated naive surprise signal. The compensated surprise score $\mathcal{S}_{comp, t}$ is calculated using the exact same hybrid loss (Eq.~\ref{eq:hybrid_loss}) between $\mathbf{z}_t$ and $\hat{\mathbf{z}}_{comp, t}$, and is similarly backpropagated at every step to drive continuous online adaptation. By effectively filtering out predictable reafferent noise, this compensation module mitigates false alarms induced by AUV maneuvers, ensuring the autonomous trigger remains robustly focused on pure semantic dynamics (Fig.~\ref{fig:surprise_signal_demo}, right/blue).

\subsection{Downstream Event Extraction for Review and Telemetry}
\label{sec:method_trigger_extraction}
The core output of DINO-Explorer is the ego motion compensated continuous surprise signal $\mathcal{S}_{comp, t}$. One downstream use of this signal, and the one evaluated in this paper, is to convert it into discrete event proposals for human review and telemetry escalation. We instantiate the causal proposal extraction interface in three steps.

First, we smooth the online signal with a one-sided Gaussian kernel using only current and past samples. With truncation radius $K = \max(1, \lfloor 3\sigma \rceil)$ and weights $w_k = \exp(-k^2 / 2\sigma^2)$, the causal smoother is
\begin{equation}
    \bar{\mathcal{S}}_t =
    \frac{\sum_{k=0}^{\min(t, K)} w_k \mathcal{S}_{t-k}}
         {\sum_{k=0}^{\min(t, K)} w_k}.
\end{equation}

Second, because underwater background variance is strongly non-stationary, we normalize against a trailing context window
\begin{equation}
    \Omega_t = \left\{ i : \max(T_{\text{warmup}} + 1, t - W_{\text{samp}}) \le i \le t \right\},
\end{equation}
and define the local statistics and adaptive threshold as
\begin{equation}
\begin{aligned}
    \mu_t &= \operatorname{mean}_{i \in \Omega_t} \bar{\mathcal{S}}_i, \qquad
    s_t = \operatorname{std}_{i \in \Omega_t} \bar{\mathcal{S}}_i, \\
    \tau_t &= \max(\mu_t + \alpha s_t, \tau_{\min}).
\end{aligned}
\end{equation}
where $W_{\text{samp}}$ is the sample count corresponding to the time horizon $W_{\text{sec}}$.

Third, a review proposal is emitted only at causal local maxima that also exceed the adaptive threshold:
\begin{equation}
    t \in \mathcal{T}
    \iff
    t > T_{\text{warmup}}
    \;\land\;
    \bar{\mathcal{S}}_{t-1} < \bar{\mathcal{S}}_{t}
    \ge \bar{\mathcal{S}}_{t+1}
    \;\land\;
    \bar{\mathcal{S}}_{t} > \tau_t.
\end{equation}
To prevent duplicate alarms for the same macroscopic phenomenon, accepted peaks are greedily filtered with a refractory interval $R$. Together, $(\sigma, W_{\text{sec}}, \alpha, \tau_{\min}, T_{\text{warmup}}, R)$ define a family of causal extraction policies built on the same surprise signal. In Section~\ref{sec:eval_operating_point}, we specify the operational point used for human review and all reported metrics.

\section{Evaluation Setup}

The evaluation follows the three research questions posed in the introduction: whether trigger proposals surface review-relevant events, whether motion conditioning suppresses maneuver-induced false alarms, and whether the trigger stream effectively reduces telemetry burden while perceive semantic novelty. We design the evaluation in two levels. The fixed operating point asks what the causal event extractor sends to reviewers under the threshold used for the annotation workflow. The later $\alpha$ sweep pareto front analysis asks a broader question: how the same extractor behaves as its sensitivity varies. This avoids judging the method only at a single selected threshold and provides an overview benchmark across the event-quality and telemetry-bandwidth trade-off.

\subsection{Dataset}
We evaluate DINO-Explorer on 178.2 minutes of 720p underwater footage collected during native-oyster restoration surveys at Borkum Riffgrund in the North Sea \cite{pinedametz2025mtid}. The data were recorded with a BlueROV2 \cite{bluerobotics_brov2_datasheet_2025} at speeds up to 1.6 m/s and depths up to roughly 40 m, and include sandy and coarse substrates, shell accumulations, reef structures, benthic fauna, and frequent suspended sediment.

\subsection{Surprise Event Taxonomy}
To operationalize the current human-review target, we group relevant events into three categories:
\begin{itemize}
    \item \textit{Spatial Transitions}: scene switches, frame entry/exit, and object discoveries.
    \item \textit{Environmental Events}: lighting changes and turbidity bursts that alter visibility.
    \item \textit{Animal \& AUV Behavior}: meaningful behavioral shifts rather than simple appearance changes.
\end{itemize}
This taxonomy defines the operational annotation target for the study.  We focus on events large enough to disrupt overall scene semantics and ignore steady-state background dynamics and minor local motion. Representative examples, except the Animal Behavior which is already shown in Fig.~\ref{fig:surprise_signal_demo}, are demonstrated in Appendix~A.

% Because individual surprise thresholds vary, the reported metrics are consensus-relative by design: they measure agreement with the annotator consensus under these categories.

\subsection{Human Annotation and Verification Protocol}
Defining novelty in continuous video is inherently subjective and prone to cognitive fatigue in humans (e.g., inattentional blindness during long sequences without events). To account for this bias and establish a fair human consensus ground truth, we conducted a two-phase verification study involving 15 human reviewers spanning different ages and backgrounds:

\begin{itemize}
    \item \textbf{Phase 1: Blind annotation.} Reviewers mark interval-level novelty events without access to model outputs.
    \item \textbf{Phase 2: Model-guided validation.} Reviewers assess clips centered on automated triggers $t^*$ that fall outside their Phase 1 intervals and mark each as \textit{Agree} or \textit{Reject}.
\end{itemize}

This yields two annotation-derived event sets: Phase 1 consensus events from blind annotation and model-guided confirmations from reviewer assessment of the surprise signal proposals.

\subsection{Operational Trigger Configuration}
\label{sec:eval_operating_point}
All human reviewed clips and operating-point tables use the downstream extractor from Section~\ref{sec:method_trigger_extraction}. Unless noted otherwise, both model variants use one-sided Gaussian smoothing with $\sigma=2.0$, a trailing threshold window of $W_{\text{sec}}=10.0$ s, sensitivity $\alpha=2.5$, minimum threshold $\tau_{\min}=0.005$, warmup horizon $T_{\text{warmup}}=200$ frames, and refractory interval $R=0.5$ s. These parameters are used during annotation and define a stable scene-level operating point for human review; the replay Pareto analysis then varies $\alpha$ to characterize the surrounding extraction family.

\subsection{Evaluation Metrics}
\label{sec:eval_metrics}
The metrics follow the two-level evaluation design above. The fixed operating point reports how the extracted surprise-event proposals align with reviewer consensus at the selected extractor threshold. The $\alpha$ sweep reports how the same extractor behaves across sensitivity settings, so the comparison is not tied to a single selected operating point.

\paragraph{Fixed operating point}
At the fixed operating point, Phase 1 Subject-Pool Consensus Recall (\textsc{SPCR}) is the recall-side human-reviewer consensus-retention metric: it measures how many Phase 1 category-consensus events are retained by the operating-point trigger stream. We also report a post-discovery \textsc{SPCR} for the fixed operating point. This companion value adds model-guided Phase 2 discovery confirmations. Phase 1 Trigger Confirmation Rate (\textsc{TCR}) is the precision-side proposal-confirmation metric: it measures the fraction of extracted proposals validated by the Phase 1 consensus. Post-discovery \textsc{TCR} keeps the denominator fixed as all extracted proposals at the operating point, but expands the numerator to include proposals confirmed during model-guided Phase 2 review. Discovery Rate (\textsc{DR}) measures additional annotator-validated events surfaced beyond Phase 1 blind review. False Positive Suppression Rate (\textsc{FPSR}) isolates how much ego-motion compensation reduces false alarms relative to the uncompensated predictive baseline:
\begin{equation}
    \mathrm{FPSR} = 1 - \frac{N_{comp}^{false}}{N_{uncomp}^{false}},
\end{equation}
where $N_{comp}^{false}$ and $N_{uncomp}^{false}$ are the corresponding false-alarm counts under the operating-point extraction rule.

\paragraph{$\alpha$-sweep replay benchmark}
Across the $\alpha$ sweep, validated peak F1 is the main event-proposal-quality measure. We match extracted trigger peaks one-to-one to reviewer-validated proposal peaks within the tolerance window and summarize the resulting peak precision and peak recall with
\begin{equation}
    \mathrm{F1}_{\text{peak}} =
    \frac{2\,P_{\text{peak}}\,R_{\text{peak}}}
    {P_{\text{peak}} + R_{\text{peak}}},
\end{equation}
where $P_{\text{peak}}$ and $R_{\text{peak}}$ denote matched-peak precision and recall. Phase 1 \textsc{SPCR} and Phase 1 \textsc{TCR} are reported in the sweep as diagnostic recall- and precision-side views against the blind Phase 1 consensus.

\paragraph{Telemetry bandwidth}
Bandwidth Savings Ratio (\textsc{BSR}) measures telemetry reduction relative to continuous 30 frames-per-second (FPS) streaming. The \textsc{BSR} computation assumes 1 FPS default streaming and 30 FPS inside a $\pm 3$ s window around each trigger:
\begin{equation}
    \mathrm{BSR} = 1 - \frac{N_{\text{tx}}}{N_{\text{raw}}},
\end{equation}
where $N_{\text{raw}}$ denotes the total raw-frame count and $N_{\text{tx}}$ the number of transmitted frames under this transmission rule.

\paragraph{Latent-energy retention}
Latent Energy Retention (\textsc{LER}) provides a replay-side diagnostic for the surprise-trigger stream that does not depend on human event labels. We first compute a reference semantic-change trace from the full video in the shared frozen DINOv3 latent space. Each trigger policy is then scored by the fraction of this reference trace retained by its selected telemetry windows. This keeps the annotation process and representation backbone fixed, so differences in \textsc{LER} reflect the trigger policy's ability to preserve DINOv3-semantic changes under downsampling. Given the frozen DINO latent sequence $\{\mathbf{z}_t\}_{t=1}^{T}$, let
\begin{equation}
    \boldsymbol{\mu}_t = \frac{1}{C} \sum_{k=t-C}^{t-1} \mathbf{z}_k,
    \qquad
    \delta_t =
    1 - \frac{\mathbf{z}_t^{\top}\boldsymbol{\mu}_t}{\|\mathbf{z}_t\|_2 \, \|\boldsymbol{\mu}_t\|_2},
\end{equation}
where $C$ is the rolling context length and $\delta_t=0$ when either norm is zero. We then median-center the cosine deviation and clip it to the non-negative part,
\begin{equation}
    e_t = \max\!\Bigl(\delta_t - \operatorname{median}\{\delta_j\}_{j=C}^{T},\, 0\Bigr),
\end{equation}
which defines the per-frame latent-change energy. Given the trigger-induced telemetry mask $w_t$, summing over the replay gives
\begin{equation}
    E_{\text{total}} = \sum_{t=1}^{T} e_t,
    \qquad
    E_{\text{captured}} = \sum_{t=1}^{T} w_t e_t,
\end{equation}
where $w_t=1$ inside trigger windows and $w_t=r_{\text{low}}$ elsewhere, with $r_{\text{low}}$ denoting the low-rate sampling ratio. We report
\begin{equation}
    \mathrm{LER} = \frac{E_{\text{captured}}}{E_{\text{total}}} \times 100,
\end{equation}
so a higher \textsc{LER} indicates greater retention of the reference latent-change trace in the downsampled telemetry stream.

\section{Results}
\begin{figure*}[!h]
    \centering
    \subfloat[Validated peak F1 versus telemetry budget]{\includegraphics[width=0.43\textwidth]{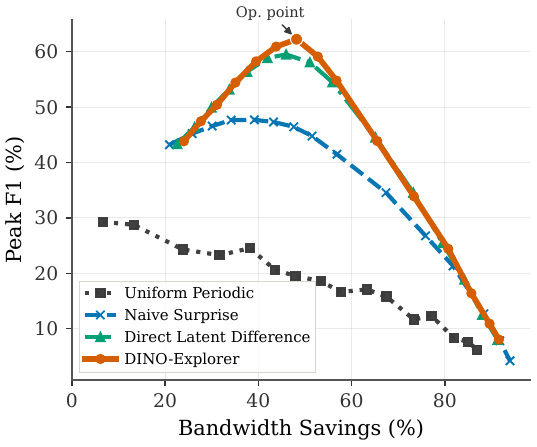}\label{fig:pareto_f1}}
    \hfill
    \subfloat[Phase 1 \textsc{SPCR} recall-side retention versus telemetry budget]{\includegraphics[width=0.43\textwidth]{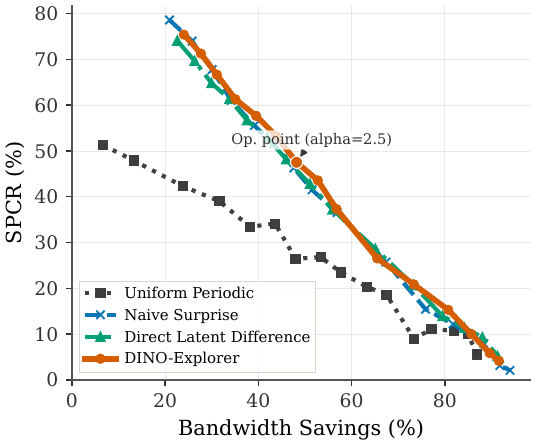}\label{fig:pareto_retrieval}}
    \\
    \vspace{0.4em}
    \subfloat[Phase 1 \textsc{TCR} precision-side confirmation versus telemetry budget]{\includegraphics[width=0.43\textwidth]{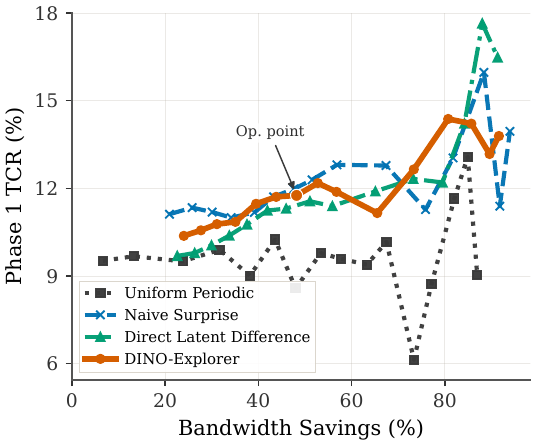}\label{fig:pareto_precision}}
    \hfill
    \subfloat[DINOv3-latent change energy retention versus telemetry budget]{\includegraphics[width=0.43\textwidth]{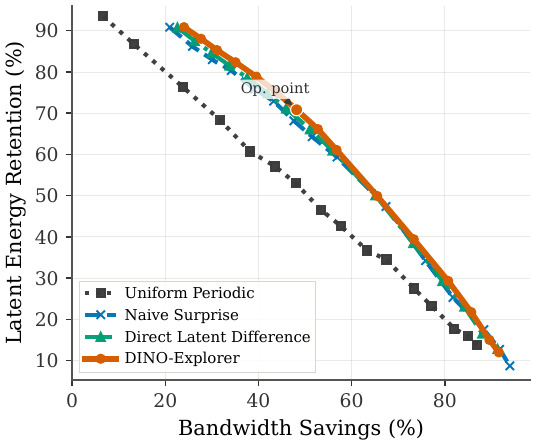}\label{fig:pareto_energy}}
    \caption{Ablation benchmark for the proposed DINO-Explorer under matched telemetry budgets. Sweeping $\alpha$ avoids tying the comparison to a single downstream event-extraction hyperparameter and shows the ablation benchmark across extraction sensitivities. Each panel is a budget-controlled frontier: at the same bandwidth saving, a higher curve indicates stronger proposal quality or better preservation of the corresponding diagnostic. The sweep compares DINO-Explorer with the uncompensated surprise signal, Direct $\Delta \mathbf{z}_t$, and uniform replay, showing that the ego-motion-compensated surprise signal improves human-reviewer-facing proposal quality while preserving the Phase 1 recall, proposal-confirmation, and DINO-latent semantic-change diagnostics needed for underwater triage.}
    \label{fig:pareto_fronts}
\end{figure*}

\begin{table}[H]
    \centering
    \caption{Operating-point consensus summary. Phase 1 (P1) uses blind same-category consensus intervals; P1+P2 additionally counts reviewer-confirmed model-guided discoveries. \textsc{FPSR} compares against the uncompensated predictive variant, and \textsc{DR} reports additional event coverage with model discovering.}
    \label{tab:consensus_summary}
    \setlength{\tabcolsep}{2pt}
    \footnotesize
    \begin{tabular}{lcccccc}
        \hline
        Setting & \multicolumn{2}{c}{\textsc{SPCR} (\%)} & \multicolumn{2}{c}{\textsc{TCR} (\%)} & \textsc{FPSR} (\%) & \textsc{DR} (\%) \\
         & P1 & P1+P2 & P1 & P1+P2 & & \\
        \hline
        $\alpha=2.5$ & 47.5 & 78.8 & 11.8 & 56.8 & 45.5 & 147.8 \\
        \hline
    \end{tabular}
\end{table}

\subsection{Reviewer-consensus validation at the operating point}

Table~\ref{tab:consensus_summary} summarizes reviewer consensus at the fixed $\alpha=2.5$ operating point used for the annotation workflow. The 47.5\% Phase 1 \textsc{SPCR} means that the trigger stream retained 47.5\% of the blind same-category consensus events. After adding reviewer-confirmed Phase 2 model-guided discoveries, post-discovery \textsc{SPCR} rises to 78.8\%, indicating broader event coverage once the model-guided review stage is included. The 11.8\% Phase 1 \textsc{TCR} means that 11.8\% of extracted proposals were confirmed by the blind Phase 1 consensus, while the 56.8\% post-discovery \textsc{TCR} means that 56.8\% of proposals were confirmed after including Phase 2 reviewer decisions. The 45.5\% \textsc{FPSR} means that ego-motion compensation reduced false alarms by 45.5\% relative to the uncompensated predictive variant under the same extraction setting. The 147.8\% \textsc{DR} means that model-guided review surfaced additional events missed during Phase 1 blind annotation; these events were proposed by the model at Phase 2 and then confirmed by majority reviewer vote in Phase 2.

\subsection{Replay pareto ablation benchmark}
Fig.~\ref{fig:pareto_fronts} evaluates the extractor over an $\alpha$ sweep, rather than only at the fixed human-review setting in Table~\ref{tab:consensus_summary}. Each point corresponds to a trigger stream produced at one sensitivity setting, and the curves show how proposal quality and diagnostic retention change as telemetry bandwidth varies. The comparison includes the compensated model, the uncompensated predictive variant, Direct $\Delta \mathbf{z}_t$ as a non-predictive frame-to-frame DINO-latent-change baseline, and Uniform periodic sampling as a content-agnostic baseline that selects telemetry windows at fixed time intervals.

Panel (a) shows that DINO-Explorer forms the strongest validated-peak-F1 frontier, with Direct-Diff as the closest simple baseline, Naive Surprise below it, and Uniform periodic sampling as the weakest reference. This indicates that DINO-Explorer improves the quality-bandwidth trade-off across extraction sensitivities: for a comparable telemetry budget, it yields trigger proposals that better align with reviewer-confirmed events than raw latent change, the uncompensated surprise signal, or plain periodic sampling.

Panel (b) provides the Phase 1 \textsc{SPCR} recall-side view against blind consensus and shows that DINO-Explorer stays near or above the recall-friendly Direct-Diff and Naive Surprise baselines, indicating that ego-motion compensation filters false alarms while retaining the core blind-consensus events annotated by reviewers.

Panel (c) is mainly a diagnostic precision-side view: it shows how often the triggered proposals are confirmed by the Phase 1 blind-consensus reference. The similar Phase 1 \textsc{TCR} curves indicate that this panel is auxiliary to the main peak-F1 frontier in Panel (a), rather than the primary source of DINO-Explorer's contribution.

Panel (d) shows that DINO-Explorer remains near or above Direct-Diff in \textsc{LER}, even though Direct-Diff is a strong latent-energy baseline because it directly tracks frame-to-frame change in the same DINOv3 latent space. DINO-Explorer also stays slightly above Naive Surprise over much of the range. This suggests that DINO-Explorer filters motion-induced residuals without losing the main DINOv3-semantic dynamics. When the compensated model exceeds the naive variants, it indicates that motion-conditioned prediction may help concentrating the retained telemetry around semantically informative dynamics beyond raw frame-to-frame latent differences.

% \begin{table}[H]
%     \centering
%     \caption{Operating-point replay comparison. Peak F1 is measured against reviewer-confirmed proposal peaks; \textsc{BSR}, \textsc{SPCR}/\textsc{TCR}, and \textsc{LER} report bandwidth, recall/precision diagnostics, and latent-energy retention.}
%     \label{tab:operating_point_replay}
%     \setlength{\tabcolsep}{2pt}
%     \scriptsize
%     \resizebox{\columnwidth}{!}{%
%     \begin{tabular}{lcccccc}
%         \hline
%         Method & Setting & \textsc{BSR} (\%) & Peak F1 (\%) & \textsc{SPCR} P1 (\%) & \textsc{TCR} P1 (\%) & \textsc{LER} (\%) \\
%         \hline
%         $\mathcal{S}_{comp}$ & $\alpha=2.5$ & 48.2 & 62.2 & 47.5 & 11.8 & 70.8 \\
%         $\mathcal{S}_{uncomp}$ & $\alpha=2.5$ & 47.6 & 46.4 & 46.3 & 12.0 & 68.2 \\
%         Direct diff ($\Delta \mathbf{z}_t$) & $\alpha=2.5$ & 45.9 & 59.5 & 48.2 & 11.3 & 71.2 \\
%         Uniform & $\Delta t = 12$ s & 48.0 & 19.5 & 26.5 & 8.6 & 53.0 \\
%         \hline
%     \end{tabular}
%     }
% \end{table}
\begin{table}[H]
    \centering
    \caption{Operating-point replay comparison. Peak F1 is measured against reviewer-confirmed proposal peaks; \textsc{BSR}, \textsc{SPCR}/\textsc{TCR}, and \textsc{LER} report bandwidth, recall/precision diagnostics, and latent-energy retention. Bold values mark the primary DINO-Explorer operating-point readouts.}
    \label{tab:operating_point_replay}
    \setlength{\tabcolsep}{2pt}
    \scriptsize
    \resizebox{\columnwidth}{!}{%
    \begin{tabular}{lcccccc}
        \hline
        Method & Setting & \textsc{BSR} (\%) & Peak F1 (\%) & \textsc{SPCR} P1 (\%) & \textsc{TCR} P1 (\%) & \textsc{LER} (\%) \\
        \hline
        $\mathcal{S}_{comp}$ & $\alpha=2.5$ & \textbf{48.2} & \textbf{62.2} & \textbf{47.5} & 11.8 & \textbf{70.8} \\
        $\mathcal{S}_{uncomp}$ & $\alpha=2.5$ & 47.6 & 46.4 & 46.3 & 12.0 & 68.2 \\
        Direct diff ($\Delta \mathbf{z}_t$) & $\alpha=2.5$ & 45.9 & 59.5 & 48.2 & 11.3 & 71.2 \\
        Uniform periodic & $\Delta t = 12$ s & 48.0 & 19.5 & 26.5 & 8.6 & 53.0 \\
        \hline
    \end{tabular}
    }
\end{table}
\subsection{Replay benchmark at the operating point}
Table~\ref{tab:operating_point_replay} reports the fixed-threshold replay comparison at the operating point highlighted in Fig.~\ref{fig:pareto_fronts}. DINO-Explorer achieves the best validated peak F1 while also providing the strongest bandwidth saving among the compared methods. Its Phase 1 \textsc{SPCR}, Phase 1 \textsc{TCR}, and \textsc{LER} remain close to the strongest diagnostic baselines.

\begin{table}[H]
    \centering
    \caption{Runtime context for asynchronous mission-time triage on an RTX A6000.}
    \label{tab:runtime_summary}
    \setlength{\tabcolsep}{6pt}
    \footnotesize
    \begin{tabular}{lc}
        \hline
        Metric summary & Value \\
        \hline
        Avg / median / p95 latency (ms) & 227.2 / 228.7 / 272.9 \\
        Average FPS & 4.40 \\
        \hline
    \end{tabular}
\end{table}

\subsection{Runtime and deployment context}
We include runtime as deployment context for asynchronous triage and as a target for future onboard optimization.

Table~\ref{tab:runtime_summary} summarizes latency and throughput from the current inference logs.

\section{Discussion and Conclusion}
We present DINO-Explorer, a motion-aware semantic predictive-coding framework that converts foundation-model latent prediction error into a continuous surprise signal for underwater robot attention in visually degraded marine environments. The evaluation on a representative downstream telemetry task supports three design claims. First, semantic prediction mismatch provides a useful abstraction for identifying scene-level novelty beyond low-level visual fluctuation. Second, conditioning prediction on ego-motion improves the separation between self-induced visual change and externally meaningful novelty. Third, the resulting surprise signal can serve as a decision variable for downstream robot-facing interfaces, as demonstrated here through event triage and selective telemetry.

\subsection{Limitations and Future Directions}
The current implementation leaves two concrete extension points. First, the inference package is validated as an asynchronous server-side pipeline on the RTX A6000; turning DINO-Explorer into a real-time onboard module will require compressing or distilling the frozen DINOv3 encoder and RAFT motion estimator and scheduling the recurrent predictor under robot compute budgets. Second, the present representation applies global average pooling over DINO latent before predicting surprise. This keeps the signal compact, but it loses local surprise detail: when surprise signal spikes, the system cannot yet identify which patch, object, or local visual transition in the frame primarily drove that surprise. Future versions should preserve patch-level or multi-scale latent maps, compute localized surprise heatmaps to provide visual cues that can indicate where in the view made the robot surprised.

These future paths keeps the current contribution centered on a measured surprise signal while making its next role explicit: from asynchronous event triage, to active exploration control, and eventually to curiosity-driven semantic world-model refinement in long-horizon robot learning.

% In future work, we plan to integrate DINO-Explorer into the continuous control loop of an onboard AUV agent to enable real-time novelty detection and robotic behavior. We plan to embed the intrinsic surprise signal as an intrinsic reward in a goal-conditioned reinforcement learning policy or as a cost term in model predictive control, enabling the AUV to actively steer toward novel phenomena while respecting safety constraints. This closed-loop architecture will enable the AUV to autonomously adapt its trajectory to unexpected events rather than following pre-defined paths. Additionally, the surprise signal, when shared in a multi-agent context, could be used to coordinate divergent exploration and avoid redundancy. Furthermore, the surprise signal can trigger model updates when detecting meaningful environmental changes, thereby promoting continual learning. Ultimately, the presented integrations will leverage DINO-Explorear's potential as a core component of autonomous underwater robotic exploration, turning passive novelty detection into active, control-aware scientific discovery.
\FloatBarrier

\section*{ACKNOWLEDGMENTS}
This work was funded by the German Federal Ministry for the Environment, Climate Action, Nature Conversation and Nuclear Safety (BMUKN) supported by the ZUG under grants 67KIA4036A and 67KIA4036C, and partially supported by the German Federal Ministry of Research, Technology and Space (BMFTR) under the Robotics Institute Germany (RIG) under grant 16ME1010.
The authors would like to thank Nael Jaber and Yi-Ling Liu for their valuable feedback and discussion on this manuscript. 

\bibliographystyle{ieeetr}
\bibliography{reference}
\appendices
\section{Qualitative Examples of Surprise Events}
\label{sec:appendix_taxonomy}

\begin{figure*}[p]
    \centering
    \subfloat[Habitat transition: semantic surprise rises as sandy substrate gives way to reef structure.]{\includegraphics[width=0.74\textwidth]{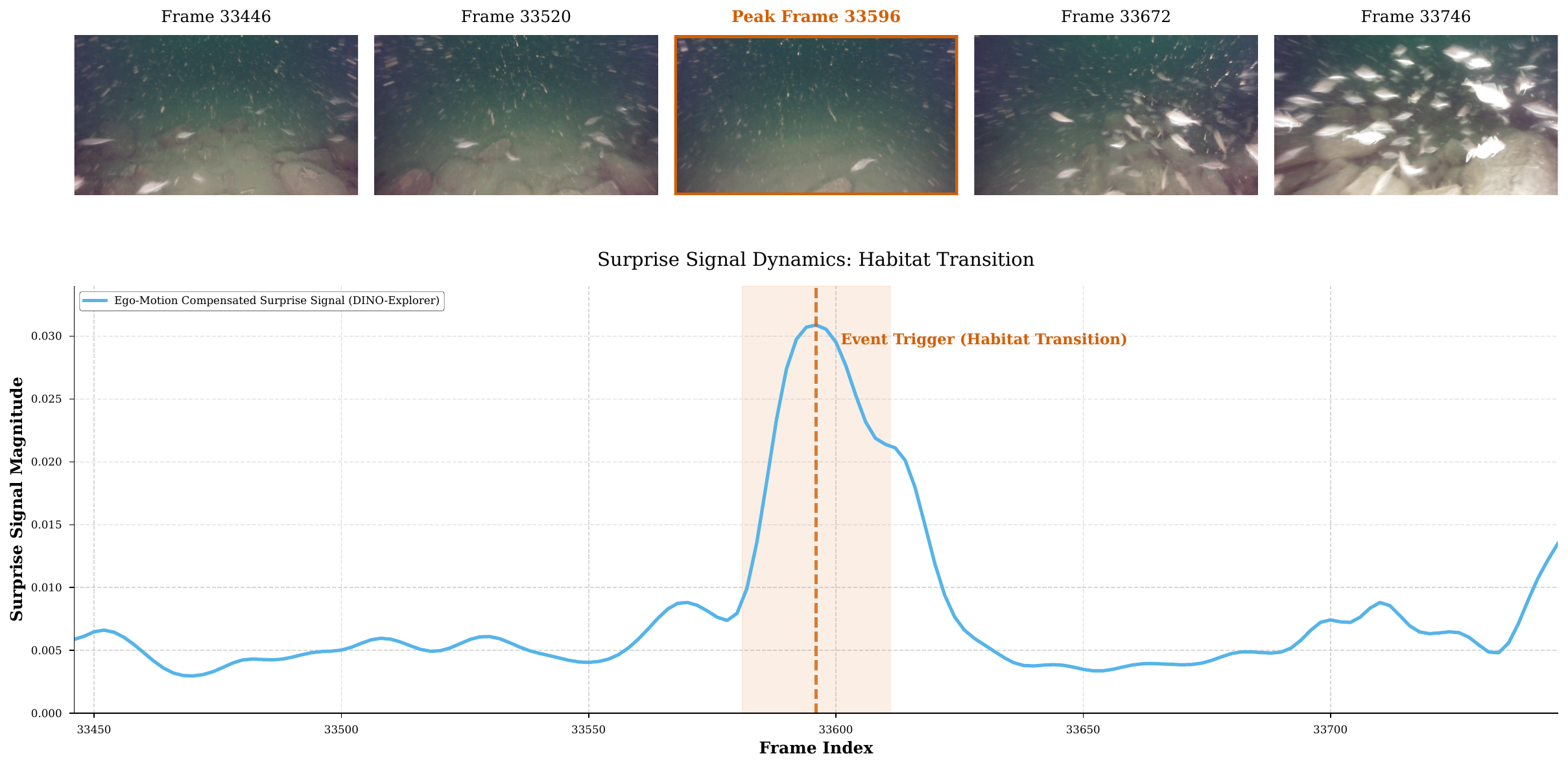}\label{fig:app_habitat}}
    \\
    \vspace{0pt}
    \subfloat[Turbidity phenomenon: a sediment plume abruptly changes scene clarity.]{\includegraphics[width=0.72\textwidth]{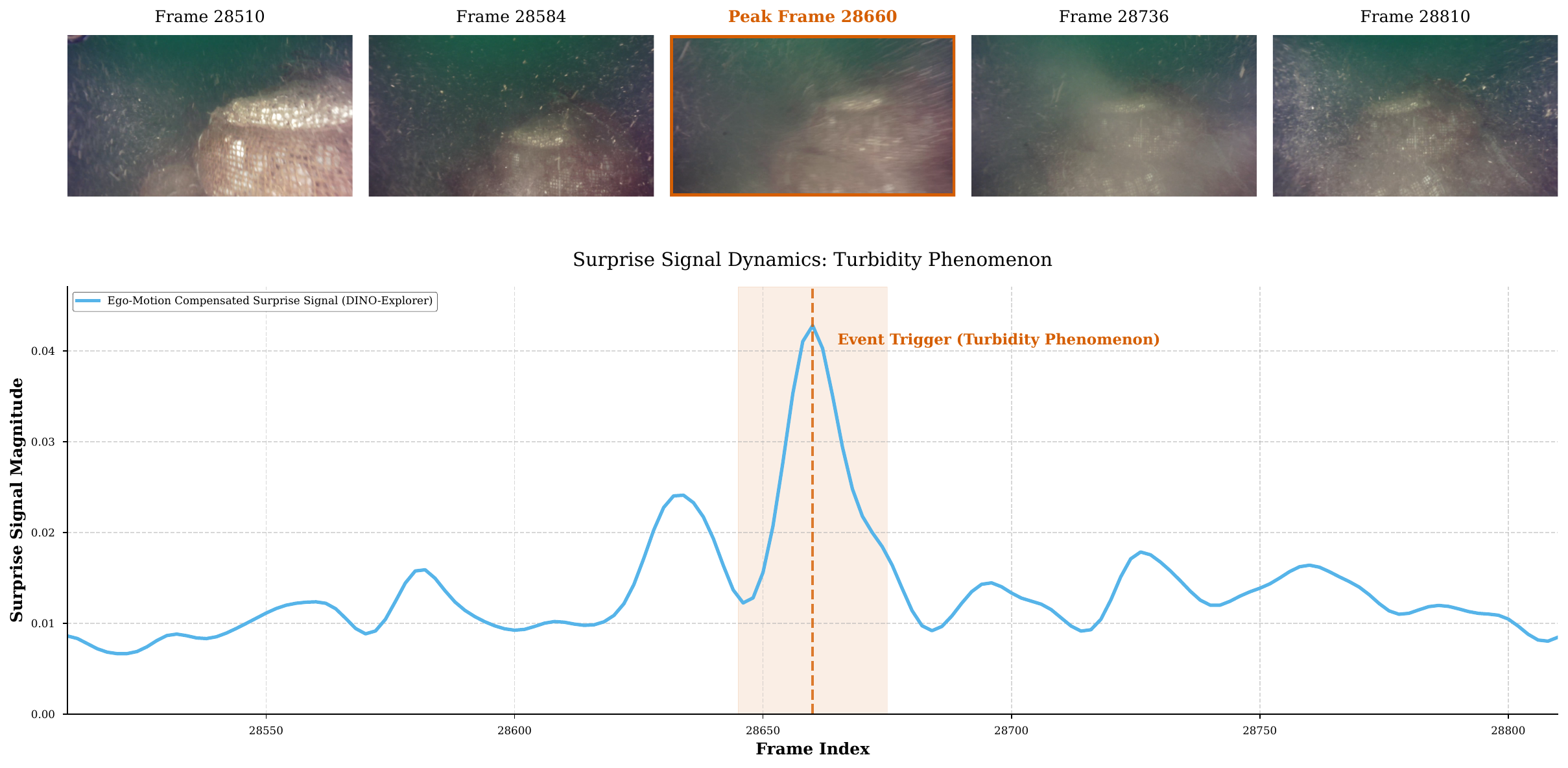}\label{fig:app_turbidity}}
    \\

    \vspace{0pt}
    \subfloat[Illumination variance: ambient light or camera exposure shifts the global appearance.]{\includegraphics[width=0.72\textwidth]{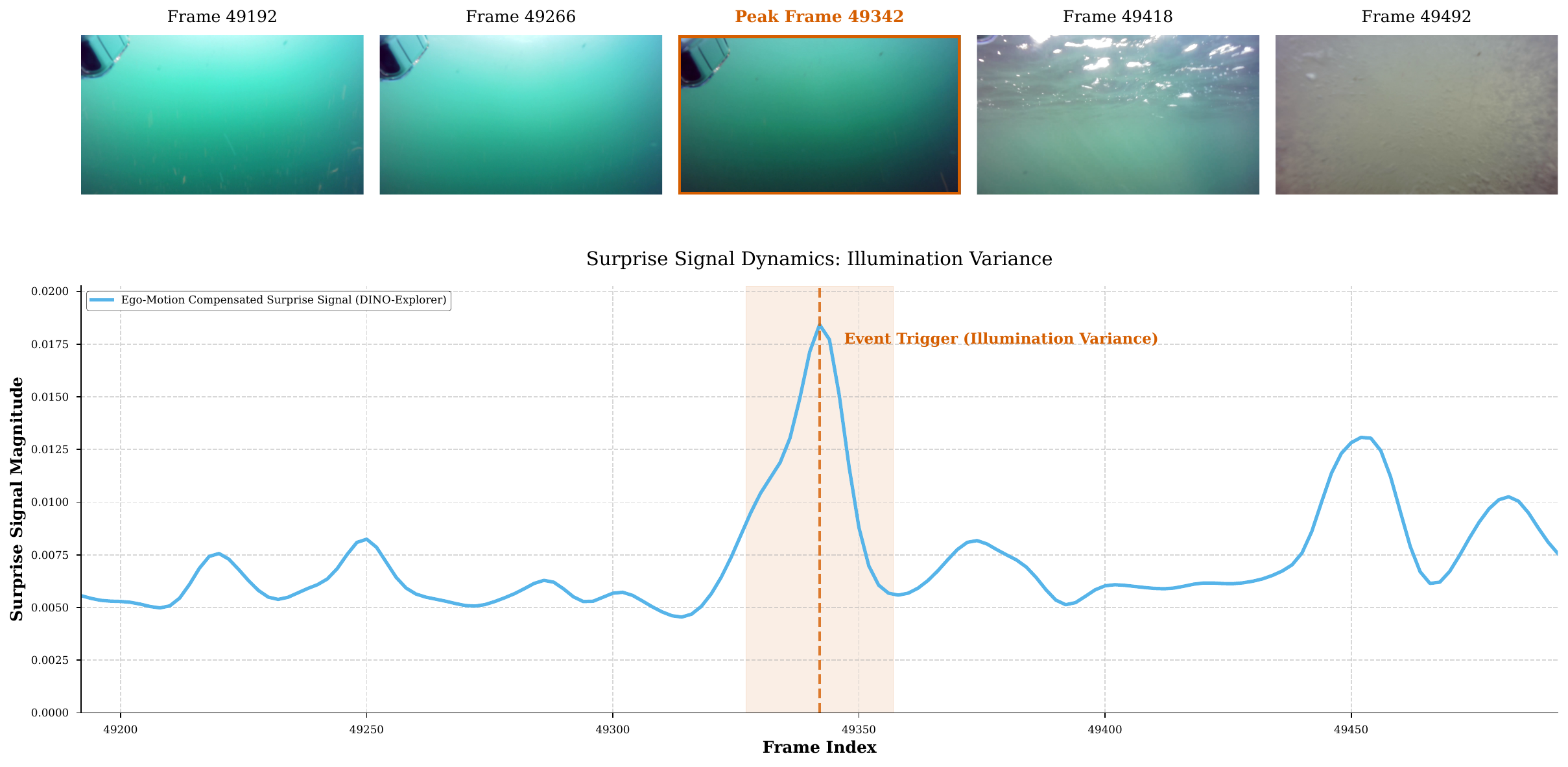}\label{fig:app_illumination}}
    \caption{Qualitative surprise-event examples. Rows show the compensated surprise trace for a habitat transition, turbidity plume, and illumination shift, illustrating low response to predictable underwater noise and peaks on event-level semantic changes.}
    \label{fig:app_env_noise}
\end{figure*}

This appendix provides qualitative examples for three non-biological surprise categories: habitat transitions, turbidity bursts, and illumination changes. Each case shows five frames with the compensated surprise signal ($\mathcal{S}_{comp}$), illustrating low response to predictable underwater noise and strong peaks for event-level changes such as reef transition (Fig.~\ref{fig:app_habitat}), sediment plume (Fig.~\ref{fig:app_turbidity}), and illumination shift (Fig.~\ref{fig:app_illumination}).

\clearpage

\end{document}